\documentclass[onecolumn]{wlscirep}
\usepackage[utf8]{inputenc}
\usepackage[T1]{fontenc}

\usepackage{float}
\restylefloat{figure}
\usepackage{hhline}

\title{Generation of microbial colonies dataset with deep learning style transfer}

\author[1,2]{Jaros\l{}aw Paw\l{}owski}
\author[1,2]{Sylwia Majchrowska}
\author[1]{Tomasz Golan}
\affil[1]{NeuroSYS, Rybacka 7, 53-656 Wrocław, Poland}
\affil[2]{Faculty of Fundamental Problems of Technology, Wroclaw University of Science and Technology,\newline Wybrze\.{z}e S. Wyspia\'{n}skiego 27, 50-372 Wroc\l{}aw, Poland}
\affil[*]{j.pawlowski@neurosys.com, s.majchrowska@neurosys.com, t.golan@neurosys.com}


\begin{abstract}
We introduce an effective strategy to generate an annotated synthetic dataset of  microbiological images of Petri dishes that can be used to train deep learning models in a fully supervised fashion. The developed generator employs traditional computer vision algorithms together with a neural style transfer method for data augmentation. We show that the method is able to synthesize a dataset of realistic looking images that can be used to train a neural network model capable of localising, segmenting, and classifying five different microbial species.
Our method requires significantly fewer resources to obtain a useful dataset than collecting and labeling a whole large set of real images with annotations. 
We show that starting with only 100 real images, we can generate data to train a detector that achieves comparable results (detection mAP~$=0.416$, and counting MAE~$=4.49$) to the same detector but trained on a real, several dozen times bigger dataset (mAP~$=0.520$, MAE~$=4.31$), containing over 7k images.
We prove the usefulness of the method in microbe detection and segmentation, but we expect that it is general and flexible and can also be applicable in other domains of science and industry to detect various objects.
\end{abstract}
\begin{document}

\flushbottom
\maketitle
\thispagestyle{empty}

\section*{Introduction}

Deep learning (DL) has recently achieved outstanding  results in several disciplines, including medicine, microbiology, and bioinformatics. DL-based automatic extraction of information from images has led to unprecedented progress in many fields. Neural networks were applied in numerous medical tasks, most extensively in radiology and pathology specialties~\cite{LEVINE2019157,NMI1,NMI2,CR1,CR2, SR1, SR2, SR3}. Moreover, their performance is frequently comparable, or even better than in the case of human experts. Furthermore, it is possible that DL algorithms could be used to extract data that would not be apparent in human analysis~\cite{alphafold}. The use of neural networks also makes it possible to automate industrial processes such as counting microbial colonies on Petri dishes, which is an important step in the microbiological laboratory to evaluate the cleanliness of the samples. Traditionally, the counting task is done manually or semi-automatically\cite{semiautomat,otsu}, but recent studies suggest that DL-based methodology will accelerate the process~\cite{agar, beznik2020deep, Jiang:20, juhas, dnn}.

% I dwa zdania o detektoarch 2-stage
To evaluate the number of microbial colonies grown on a Petri dish, neural network models called detectors may be applied~\cite{agar}. The most representative group of detectors are two-stage models. These architectures are characterised by high localization and classification accuracy. The concept is to separately resolve the localization and classification tasks in two stages. During the first stage, regions of interest are generated, and then in the second stage, a bounding box regression is made, and the achieved region proposals are assigned to the appropriate class with some level of confidence~\cite{Jiao_2019}. The baseline two-stage architecture is Faster R-CNN~\cite{bib:Faster2015}. The Mask R-CNN~\cite{bib:Mask2017} extends Faster R-CNN by adding a branch for predicting an object mask in parallel with the existing branch for bounding box recognition. However, object instance segmentation needs annotation at a pixel level, which makes the labeling process even more challenging. Estimation of the number of microbial colonies may also be based on density map estimation~\cite{Jiang:20}, similarly to the case of human crowd counting~\cite{ZhuZhang21}. In this approach, the annotations for each object are in the form of a density map. Algorithms based on a Convolutional Neural Network (CNN) learn the mapping between the extracted features and their object density maps, to indirectly count objects by integration of the predicted map. In turn, in this type of research UNet~\cite{bib:unet}-based models are used.

However, despite the broad possibilities and the recent interest in the topic~\cite{agar, beznik2020deep, Jiang:20}, the application of DL-based methods in medical, pharmaceutical, or food industries still faces some challenges. DL algorithms are quite efficient when learning from large amounts of data, whose collection and annotations require a lot of time and financial support. Moreover, the labelling process itself can be challenging. In our case, the problem of identifying a separate colony could be hard even for trained professionals, because some microbial colonies tend to agglomerate and overlap, thus becoming indistinguishable. On top of that, the identified colony can be atypical or deformed, and the image itself is influenced by various uncontrolled natural conditions. Object diversity requires well-annotated data whose distribution should match the target industrial use case. 

% Neural style transfer review (tez dwa zdania nt. cyclegany czy conditionalgany (pixtopix))

\subsection*{Related works}
DL can also resolve this issue by using techniques to generate synthetic data confusingly similar to real samples.
The idea is to train a model on a synthetically generated dataset with the intention of \textit{transfer learning} to real data.
The most popular models to generate synthetic data are Generative Adversarial Networks (GANs)~\cite{bib:gan}, which are composed of two neural networks---a generator to generate new samples, and a discriminator to recognize fake samples from the real one. Moreover, a CycleGAN~\cite{bib:cyclegan} composed of two GANs, or conditional GAN\cite{congan}, is useful in the case of image-to-image translation.

Generative neural models are becoming more widely used in medicine~\cite{ganmed0,nifty}. 
GANs are used for generation of high quality radiology images, e.g. mammograms for radiology education~\cite{mamgan}, or realistic chest X-ray images~\cite{chestgan}. In dermatology, for automated skin cancer classification~\cite{skingan}, or to build a general skin condition classifier~\cite{dermgan}. In ophthalmology, for eye's retinal image synthesis~\cite{retina1,retina2}.
Image-to-image translation can be also applied in dermatology for melanoma detection~\cite{melanomagan},
or to translate magnetic resonance images from computed tomography ones~\cite{mrigan}.
On the other hand, there is notable work in microbiology, related to our, on generating a dataset of blood plates on agar using GANs, then used for training a neural network model for semantic segmentation of bacterial colonies~\cite{bib:Andreini2020}.
However, GANs, like other deep learning models, need a lot of real data for the training process.

It is also worth noting that there are very few useful datasets containing images of bacterial colonies that can be used to train DL models. The biggest ones are MicrobIA~\cite{microbia}, AGAR~\cite{agar}, and DIBaS~\cite{dibas}. Collecting a large dataset is a great effort, not only to gather the images but also to label them. Therefore, creating synthetic datasets is a good alternative.

%Characterisation of bacterial colonies grow, or their detection and classification can be also performed using more advanced optical methods than using simple static RGB images. Time-lapse imaging of colony growth on dishes\cite{lapse,dnn} or its microscopic view\cite{micro}, or light polarization decomposition\cite{polar} can give more insight into complete colony characterization or enable its early detection\cite{dnn,micro,lapse}.

% W SR jest troche pracek z szalkami wiec tam zacznijmy
% potencjalni referee 
% (1) Aydogan Ozcan
% Electrical and Computer Engineering Department, University of California, Los Angeles, CA, 90095, USA
% ozcan@ucla.edu
% (2) Pezhman Sasanpour
% Department of Medical Physics and Biomedical Engineering, School of Medicine, Shahid Beheshti University of Medical Sciences, Tehran, Iran
% pesasanpour@sbmu.ac.ir
% (3) Clément Vulin
% Department of Infectious Diseases and Hospital Epidemiology, University Hospital Zurich, University of Zurich, Zurich, Switzerland
% clement.vulin@usz.ch
% (4) Jae Hee Jung
% Department of Electrical Engineering, California Institute of Technology, Pasadena, CA 91125, USA
% jhjung@caltech.edu
% (5) Morteza Mahmoudi
%Department of Anesthesiology, Brigham and Women’s Hospital, Harvard Medical School, Boston, Massachusetts, 02115, United States
% mmahmoudi@bwh.harvard.edu
% Nature Computational Methods -- pierwszy wybór
% IEEE transactions on medical imaging - moze tu pozniej wyslac
% Computer Methods and Programs in Biomedicine - wysyłka w kolejnym kroku

\subsection*{Synthetic microbial colonies dataset}
In our study, we develop another strategy to generate annotated synthetic datasets without the need for a lot of input resources. We propose a method combining traditional computer vision and DL-based style transfer concepts for generating synthetic images of Petri dishes. 
We utilise the neural style transfer technique to change the style of an image in one domain (synthetic) to the style of an image in another domain (real), and thus improve the realness of the generated data. Photorealistic image stylization does not require much resources and typically uses a single style image during the stylization process~\cite{li:19}.
The idea was introduced by Gatys\cite{Gatys:16} and further developed by Fei-Fei and others\cite{johnson:16,zhang:17,li:17,yoo:19}.
%The main goal of these types of experiments is that after applying style transfer, the image content should look like a real shot. The original structure and image's details should also stay untouched, without additional irregular distortions and reduction of image size.

We used the \textit{higher-resolution} subset of the recently introduced AGAR microbial colony dataset~\cite{agar} to conduct a synthetic dataset generation experiment. We randomly selected 100 annotated images from this subset and used them to feed the colony extraction part. Additionally, 10 empty dishes were taken to serve as a background on which colonies were deposited.
Using the extracted colonies, we generated a big synthetic dataset of Petri dishes augmented using style transfer with 20 different styles. For diverse styles, we used 20 fragments with different lighting conditions, selected from these 100 input images.
Each generated image contains a selected number of colonies (corresponding to real data to be mimicked) and is of size $512\times512$~pixels, which corresponds to the size of fragments (patches) used for training colony detectors when evaluating AGAR dataset\cite{agar}. To cover the whole dish we would need approximately $8\times8$ patches which gives a similar resolution for the whole dish image as in the AGAR samples with $4000\times4000$ pixels (\textit{higher-resolution} subset). Note, however, that during the generation process we synthesize individual patches, not the entire dish at once, but after simply adjusting the method, we could also generate the whole dish.
Finally, a deep learning detector was trained entirely on this synthetic dataset and evaluated on the test part of the \textit{higher-resolution} AGAR subset. This way, the obtained results are comparable with the ones from [\citeonline{agar}], obtained for the same DL detector architecture, but trained entirely on the real dataset, i.e. training part of the \textit{higher-resolution} subset containing 7360 images.

\section*{Methods}
In this section we present a detailed description of a method for generation of synthetic images with microbial colonies. Then we provide an overview of the neural network models used to: make the generated images more realistic, and to implement the colony detector.
Our source code with the python implementation of the generation  framework is publicly available at~[\citeonline{soft}].
\begin{figure}[t!]
\centering
\includegraphics[width=.9\linewidth]{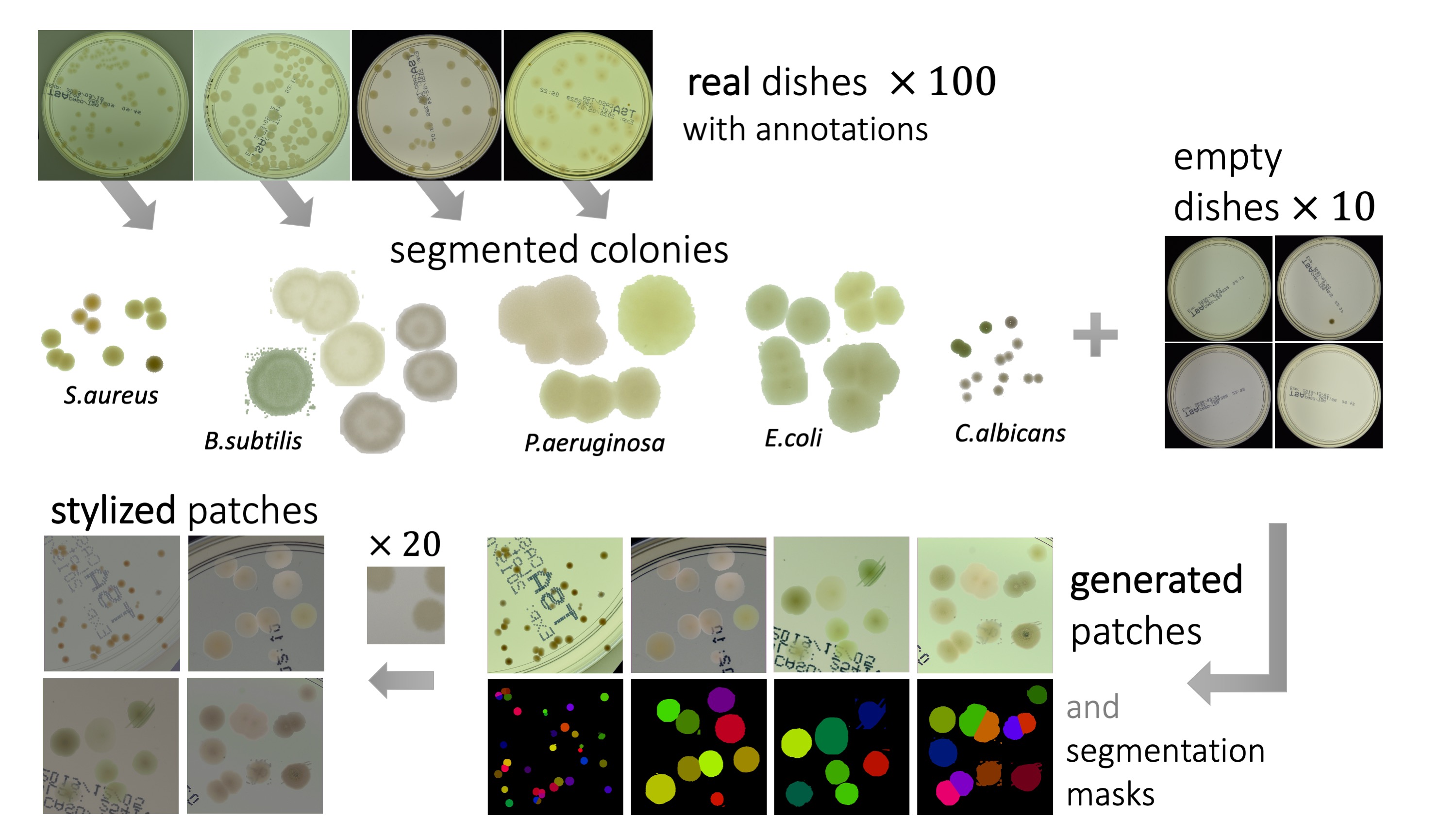}
\caption{Scheme for synthetic microbial dataset generation. Microbial colonies are segmented from real images using traditional computer vision algorithms, and then randomly arranged on fragments of an empty dish giving synthetic patches with precise annotations. To improve the realism of generated data, patches are then stylized using a neural style transfer method.}
\label{fig:schem}
\end{figure}
Let us start with a general scheme of the introduced method. In Fig.~\ref{fig:schem}, our strategy to generate a synthetic dataset is presented. 
The method consists of three subsequent steps.
We start with labeled real images of Petri dishes and perform colony segmentation using traditional computer vision algorithms including proper filtering, thresholding, and energy-based segmentation. To get a balanced working dataset, we randomly select 20 images for each of the 5 microbial species (giving 100 images in total) from the \textit{higher-resolution} subset of the AGAR dataset\cite{agar}. 

In the second step, segmented colonies and clusters of colonies are randomly arranged on fragments of an empty Petri dish (we call them patches). We select a random fragment of one of the 10 images of a real empty dish. We repeat this step many times, placing subsequent clusters in random places making sure they do not overlap. Simultaneously, we store the position of each colony from the cluster placed on the patch and its segmentation mask, creating a dictionary of annotations for that patch.

Finally, in the third step, we apply the augmentation method via deep learning style transfer. We transfer the style of a given raw patch to one of the selected real images that serve as style carriers. We select 20 real fragments with very different lighting conditions serving as style carriers to increase the diversity of the generated patches.

\subsubsection*{Colony segmentation}
At first we read the annotations---each stored as 4 numbers $(x,y,width,height)$ defining a \textit{bounding box} that represents a single labelled colony on a given image. Using this information, we check which colonies overlap by more than 0.01 part of its area and, this way, we build an \textit{adjacency matrix} between colonies. This step is important because colonies naturally overlap in highly populated dishes. Consequently, to capture the geometrical details of their mutual shapes, we want to segment whole clusters of overlapping colonies---not just individual ones. Using the adjacency matrix and breadth-first search algorithm for the graph (representing connections between colonies), we obtain connected clusters of bacterial colonies (clearly, some clusters contain a single colony) that we segment in the next step.

In this step, we cut off the rectangular fragment of the image which contains a single cluster (and repeat this step for every cluster), and add a binary \textit{alpha channel} $m_{bx}$ with zero values on areas not covered by any of the bounding boxes from the cluster. We call it boxes' mask, and obviously $m_{bx}$ has the same size as the rectangular fragment. We are now ready to perform a segmentation procedure to remove the colony background and -- consequently -- refine the alpha channel for that fragment. Examples of colonies with removed background ready to be arranged on an empty dish are presented in Fig.~\ref{fig:schem}.

\begin{figure}[h]
\centering
\includegraphics[width=.63\linewidth]{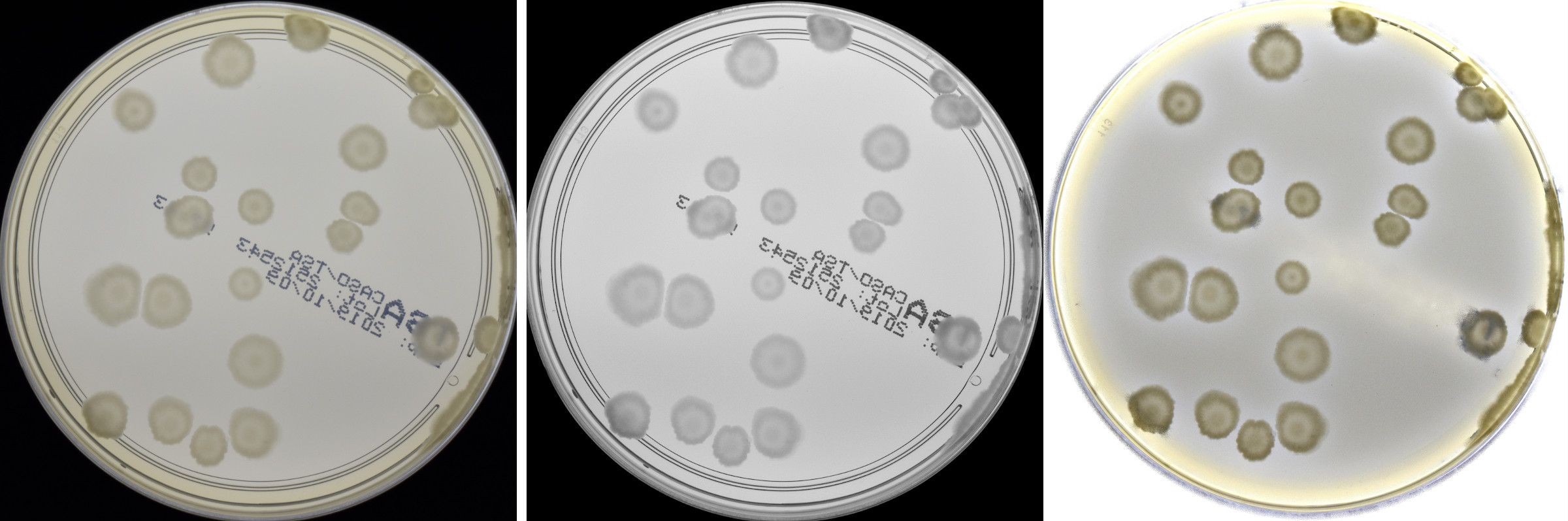}
\caption{Removal of artifacts detected by thresholding in the CIELab colorspace (\textit{luminance} channel presented in the middle), and replaced by a nearest valid pixel found by a random walk.}
\label{fig:gen_masks}
\end{figure}

The segmentation procedure consists of multiple substeps using computer vision methods. Such methods have already been used to detect and segment microbial objects\cite{semiautomat,otsu,cv,ferr,lapse}.
The first step consists of filtering out unwanted artifacts. At first, we apply the unsharp mask filter and convert the given fragment to the CIELab colorspace. Secondly, we implement simple removal of dark objects (i.e., contamination and characteristic text labels across the dish substrate): they are detected via the \textit{luminance} and \textit{b}-value thresholding\cite{semiautomat} in the CIELab colorspace. Then, we dilate the mask $m_d$ for better coverage of artifacts (dark regions). Finally, each pixel in the detected dark region is replaced with the nearest valid (i.e., not belonging to the dark region mask $m_d$) pixel found by a random walk algorithm. An example of the method operation is shown in Fig.~\ref{fig:gen_masks}.
Such an approach successfully removes the above-mentioned artifacts, but unfortunately random walk introduces a lot of speckle noise to a given fragment. Therefore, in the last step we apply denoising using non-local means\cite{nl_den} to remove such speckle noise.

\begin{figure}[h]
\centering
\includegraphics[width=.5\linewidth]{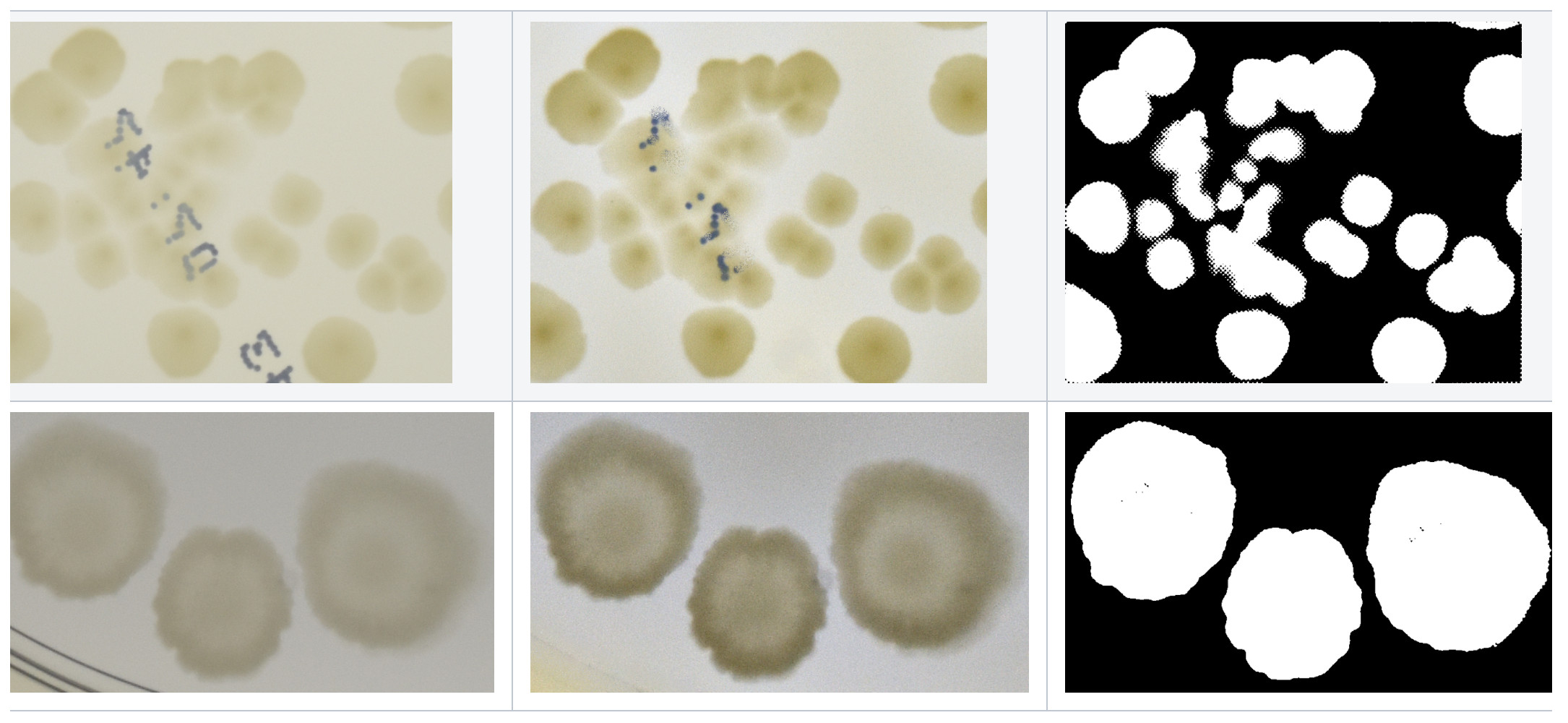}
\caption{Example patch segmentation (binary masks $m_s$ on the right) using the Chan-Vese algorithm.}
\label{fig:patch_masks}
\end{figure}

Now we are ready to perform colony cluster segmentation. 
We use the powerful Chan-Vese algorithm\cite{chan1,chan2} based on a signal energy minimization provided by the scikit-image library~\cite{scikit}.
Then add some margins to the resulting segmentation mask $m_s$ to be sure that we segment whole colonies together with their edges---see examples in Fig.~\ref{fig:patch_masks}. To get better future blending with the empty dish background, we create an additional mask $m_b$ with values for each pixel (in a range of $[0:255]$) proportional to its distance from average background color (i.e. outside the segmentation mask $m_s$) in the CIELab colorspace. Such a mask has lower values in areas where the pixel color is similar to the patch background average color. The idea in this step is to increase opacity (lower alpha channel) in areas that are similar to the background and contain a lower amount of interesting information, i.e., we assume that colonies are  different from their background.   
This step can significantly improve the realism of the visual look of the generated patches. 
Next, this blending mask multiplies the segmentation and box masks, and using Hadamard product of $m_{bx} \circ m_s \circ m_b$, we obtain the final alpha channel. Finally, we use it to remove the colony background.

\subsubsection*{Patch generation}
After extracting numerous clusters of colonies for every 5 types of microbes, we are ready to generate a synthetic dataset. To generate a single patch, we select one of the 10 empty dishes, randomly rotate it, and take its $512\times512$ sized patch. Moreover, one of the microbe types is selected. Then, we take a random colony cluster of that type, rotate and flip it randomly, and put it in a random place on the patch. The position of each placed colony (i.e., bounding box) and its segmentation mask is stored. Next, we repeat this step many times placing subsequent clusters in random places but, at the same time, making sure they do not overlap. We also select the number of colonies placed on the patch using the exponential probability distribution $\lambda e^{-\lambda x}$, with $1/\lambda=10$ mean. This corresponds to the distribution of the number of colonies on the patch in AGAR dataset. Note, however, that in case of larger colonies typically fewer colonies will fall into a single patch. 
We did not observe such phenomena in our experiments, but in some cases, to avoid biasing different classes by the size of colonies, it may be also helpful to provide an additional augmentation technique by varying (scaling) the size of placed colonies during the generation.
We repeat the whole patch-generating process and get a dataset of 50k patches. All subsequent steps of the patch generation method together with the number of parameters that need to be tuned are listed in Table~\ref{tab:parameters}.

\begin{table}[h]
\centering
\begin{tabular}{|l|c|}
\hline
\rowcolor[HTML]{EFEFEF} 
subsequent steps & no of parameters to be tuned \\ \hline
colonies clustering   & 1 \\
\hline
unsharp mask filtering  & 2  \\
\hline
removal of artifacts:    & \\
- thresholding in CIELab space & 2\\
- mask dilation & 1\\
- speckle noise cancellation & 2\\
\hline
Chan-Vese segmentation & 2 \\
- mask dilation & 1 \\
\hline
blending mask & 1 \\
\hline
neural style transfer & 1  \\ 
\hline
total & \textbf{13} \\
\hline
\end{tabular}
\caption{\label{tab:parameters} Number of parameters (that take non-default values) for each step in the patch generation method.}
\end{table}

\subsubsection*{Neural style transfer}
\begin{figure}[ht]
\centering
\includegraphics[width=.9\linewidth]{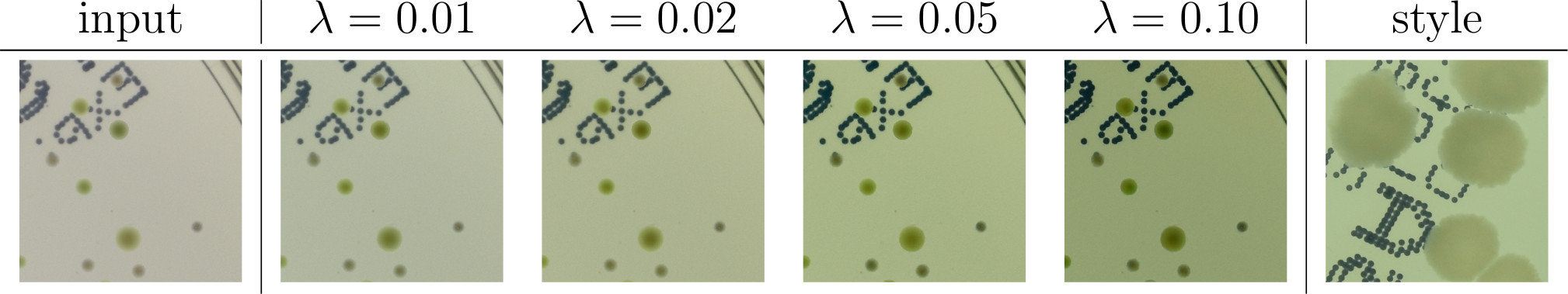}
\caption{Image style transfer for different stylization weights. Style (right) is transferred to input content image (left) with different stylization amount (middle) tuned by $\lambda$ parameter.}
\label{fig:stylization}
\end{figure}

Having described how to generate raw patches, we move on to neural style transfer. We utilize the stylization method described in [\citeonline{li:19}] with architecture similar to this used in seminal work [\citeonline{johnson:16}] but with HRNet\cite{hrnet} high-resolution representation network instead of a standard convolutional encoder-decoder architecture.
We chose this approach after several trials because it gives the most realistic stylization of our raw Petri dish images without introducing unwanted artifacts, which was the case for other tested methods, e.g, the original one introduced in [\citeonline{Gatys:16}].
We base our style transfer implementation on the code repository provided in [\citeonline{li:19}].
The architecture consists of two deep networks: an image generation network---HRNet and pretrained VGG19\cite{vgg} used to calculate content and style reconstruction losses. 
Deep convolutional neural network VGG19 is used because we are not interested in exactly matching the pixels of output $y$ and style $y_s$ or content $y_c$ images; instead we force them to have similar feature representations and compare the VGG-activations between output and style or content. The resulting feature maps are combined via Gram matrix\cite{Gatys:16} $G$, and the weighted differences
\begin{equation}
\mathcal{L}(y,\,y_c,y_s)=(1-\lambda)\Vert G(y)-G(y_c)\Vert+\lambda\Vert G(y)-G(y_s)\Vert
\end{equation} 
give the loss function $\mathcal{L}$ with $\lambda$-weight. 
Increasing $\lambda$ introduces more style to the content image as presented in Fig.~\ref{fig:stylization}.
The image generation network is trained using Stochastic Gradient Descent to minimize $\mathcal{L}$.

Stylization is the longest step in the generation method. To speed up the stylization procedure, a bigger image of size $1024\times1024$ containing 4 patches was stylized using a single style at once. Training such a one-shot stylization model lasts about 30 seconds. Moreover, in the subsequent stylization iterations, the model initial weights were taken from the previous ones, and this considerably speeds up the convergence of the stylization training.

\subsubsection*{Neural object detector}
We trained two widely used deep learning object detectors from R-CNN family: Faster~\cite{bib:Faster2015} and Cascade~\cite{bib:Cascade2018} using synthetic data, and show the usefulness of the generation method by testing the performance of microbe detection on \textit{real} images of Petri dishes---the test part of the \textit{higher-resolution} subset of AGAR dataset.
For detecting and counting microbial colonies, we used Faster R-CNN with ResNet-50~\cite{bib:resnet2016} as a backbone, and Cascade R-CNN with HRNet~\cite{hrnet} backbone. The same architectures were used as the baseline (Faster R-CNN) and a more advanced model (Cascade R-CNN) during experiments with the real AGAR dataset~\cite{agar}, which makes them good networks to compare performance. 
Moreover, the Faster detector was extended to Mask R-CNN in case of instance segmentation experiments. 
Both used detectors are the top deep learning models with quite complex structure, e.g.  Cascade R-CNN model with HRNet has got 956 different layers in total.

We trained the networks for 20 epoches, with a batch size of 8 (for Faster) or 3 (for Cascade), and an adaptive learning rate initialized to 0.0001. During training, Stochastic Gradient Descent (SGD) method was used for optimizing the network weights. Calculations were performed on NVIDIA Titan RTX GPU, and the training process lasted about 2 (Faster) to 4 (Cascade) days depending on the neural network model used. 
We used the detectors implementation from the MMDetection~\cite{bib:mmdetection2019} toolbox. Backbones’ weights were initialized using models pre-trained on ImageNet~\cite{bib:imagenet2009}, available in the \textit{torchvision} package of the PyTorch library.

\section*{Results}
In this section we present results for the generation of synthetic images with microbial colonies, and the results for
training deep learning detectors using the generated data.

\subsection*{Generation and stylization of synthetic dataset}
\begin{figure}[bt]
\centering
\includegraphics[width=0.75\linewidth]{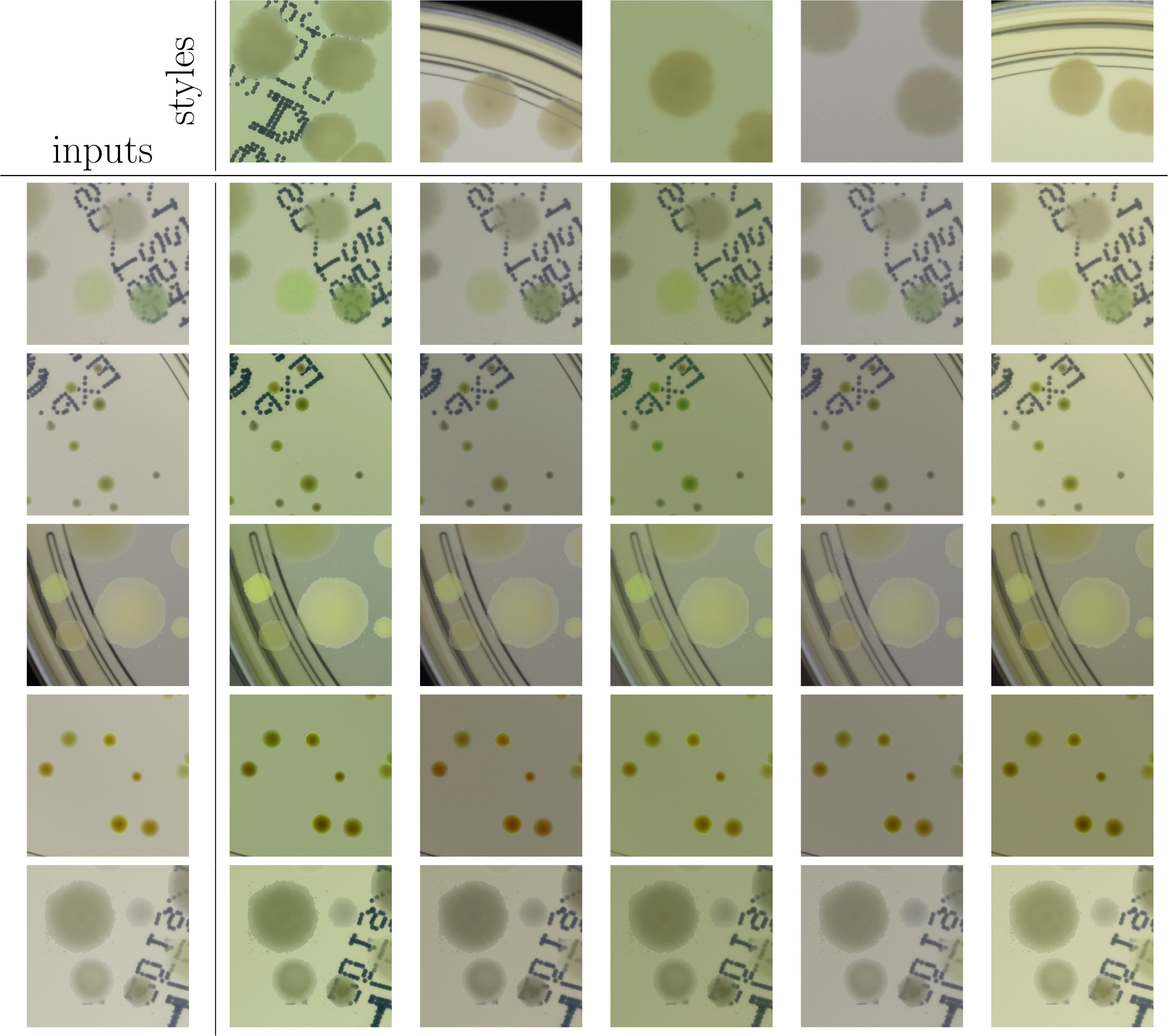}
\caption{Stylization of generated microbial data. Five synthetic patches (left) are stylized (with $\lambda=0.05$) using the style of five real images (top). Stylization improves realism of the generated images and significantly increases performance of a deep learning detector trained on this data.}
\label{fig:stylized}
\end{figure}

In the first step, we take 100 annotated real images of Petri dishes (20 for each microbe type in AGAR dataset) and perform colony segmentation using traditional computer vision algorithms.
\begin{figure}[tb]
\centering
\includegraphics[width=.7\linewidth]{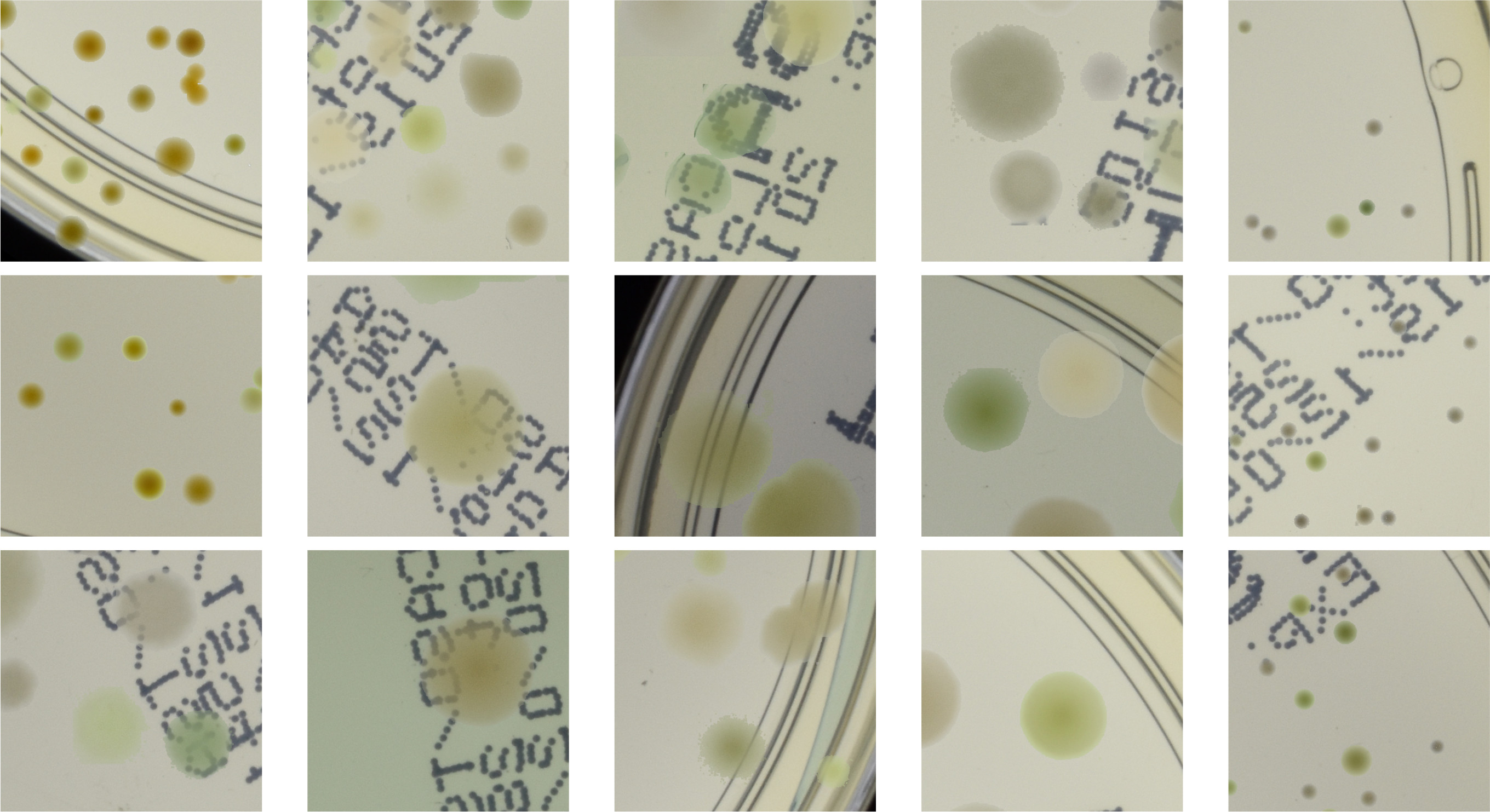}
\caption{Examples of generated synthetic patches before the stylization step. In some situations, colonies do not blend well with the background, and their color does not match the background color.}
\label{fig:gen}
\end{figure}
In the second step, segmented colonies and clusters of colonies are randomly arranged on fragments of an empty Petri dish. Examples of generated patches together with their masks denoting different colonies are shown in Fig.~\ref{fig:schem}. Further examples of generated patches with different microbe types are listed in Fig.~\ref{fig:gen}.
In the third step, we apply the augmentation method via deep learning style transfer.
Examples of stylization of five different raw synthetic patches using five different real style images are presented in Fig.~\ref{fig:stylized}.

We introduce stylizing for two visual reasons. First, it makes the generated images look more realistic---they become difficult to distinguish from real images of Petri dishes. Second, stylizing mitigates the mismatch at the edges of pasted colonies, which is an artifact that strongly impacts the detector performance---the detector would undesirably learn to detect nonrealistic edges of microbial colonies. 
However, another important factor is that the addition of stylizing, and thus augmenting using these 20 different styles, simply increased the diversity of the quite homogeneous raw dataset, and had a big impact on detector training.

Using the method, we generated 50k of \textit{raw} patches with different microbe types and their amount. To test the stylization impact, we use two different stylization strengths controlled by $\lambda$ weight---see \textit{Methods} section for further details, and apply them separately to the \textit{raw} patches. Weaker stylization with $\lambda=0.02$ gives \textit{semi-stylized} set, while stronger one with $\lambda=0.05$ gives a \textit{fully-stylized} set. We also store the \textit{raw} set with no stylization. Each of the three sets contains 50k patches which corresponds to about 65k of patches in the training part of the \textit{higher-resolution} subset of AGAR dataset.

\subsection*{Deep learning detection on real data}
The idea behind the conducted experiments is to train a neural network model using synthetic data to detect microbial colonies, and then test its performance on real images with bacterial colonies in Petri dish.

\subsubsection*{Examples of detection on real data}
\begin{figure}[bt]
\centering
\includegraphics[width=.9\linewidth]{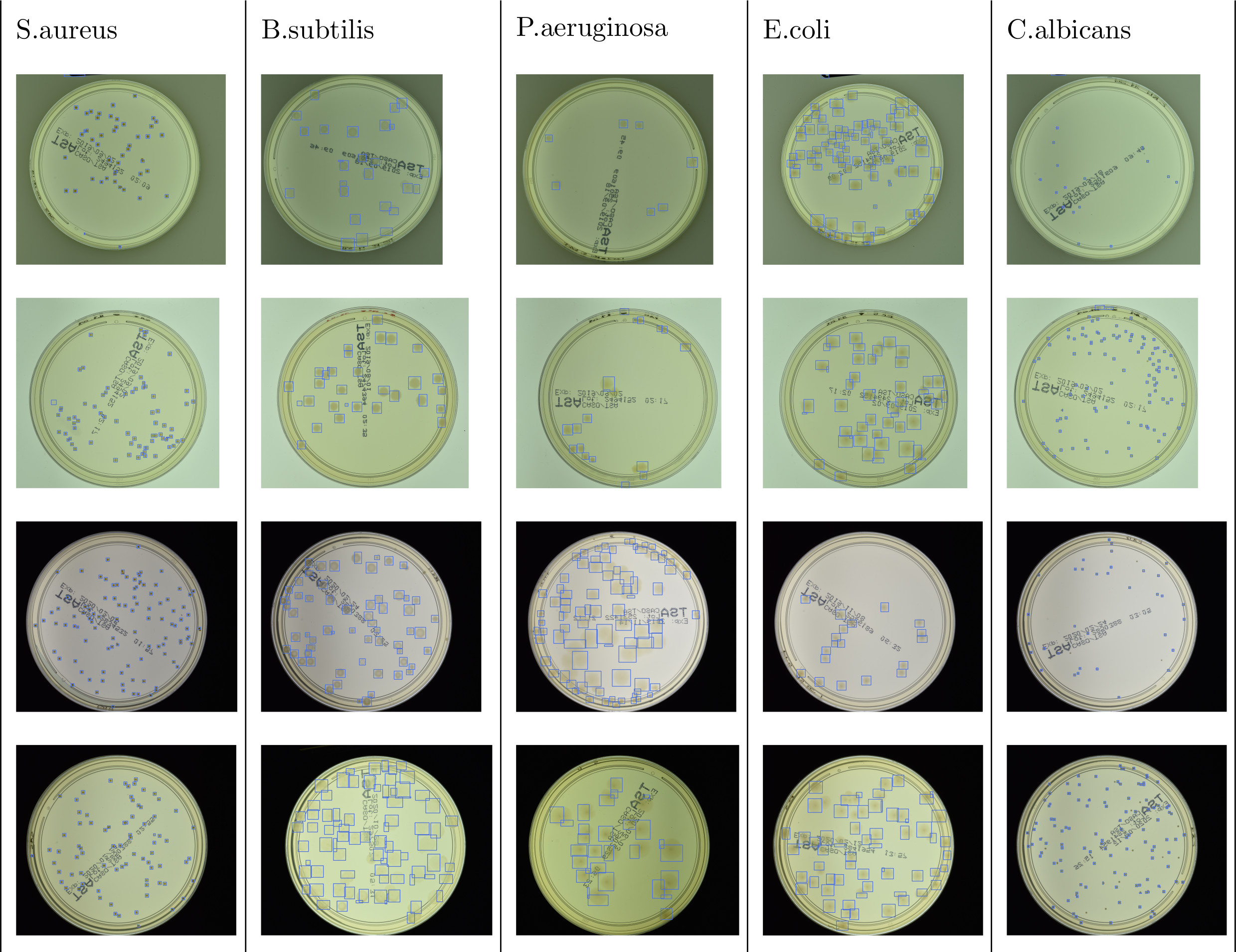}
\caption{Examples of microbial colonies detection on real data---Petri dishes from AGAR dataset. The Faster R-CNN detector correctly detects the vast majority of colonies, however some missing bounding boxes especially for crowded samples can be observed.}
\label{fig:det}
\end{figure}

We train the Faster R-CNN and Cascade R-CNN on the three synthetic sets separately. The best results are obtained for training on the \textit{fully-stylized} set---the examples of tests for real patches in case of Faster R-CNN are presented in Fig~\ref{fig:det}. Detector performs quite well, but one may notice missing bounding boxes for some colonies, especially in crowded samples where colonies frequently overlap. The worst performance occurs for blurred \textit{P.aeruginosa} microbes that form the biggest colonies; the best occurs for sharp small colonies of
\textit{S.aureus} and \textit{C.albicans} microbes. In some cases, some excessive (false positive) bounding boxes also appear.
In case of Cascade R-CNN, the results are similar with a slightly lower tendency of the detector to generate excessive bounding boxes, which result in better counting statistics---see Table~\ref{tab:res}.

\begin{figure}[bt]
\centering
\includegraphics[width=.9\linewidth]{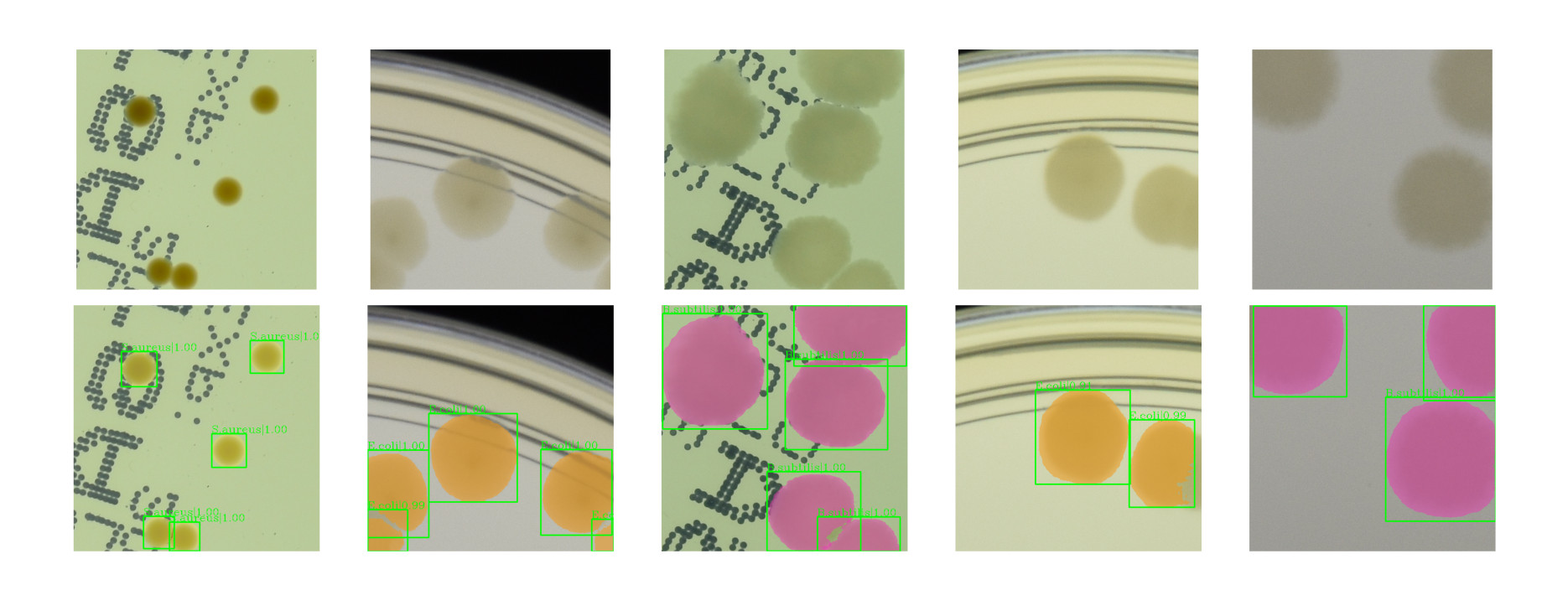}
\caption{Example results for instance segmentation of real samples in case of different microbial species.}
\label{fig:masks}
\end{figure}

Automatic instance segmentation is a problem that occurs in many biomedical applications.
During the patch generation, we also store a segmentation mask at a pixel level for each colony. This additional information can be used to train a deep learning instance segmentation model. We use the Mask R-CNN~\cite{bib:Mask2017} model which extends Faster R-CNN detector that we have already trained. 
In the study, we train the model using pretrained weights and refine colony bounding boxes obtained from the detector model.
The segmentation results for real samples are presented in Fig.~\ref{fig:masks}. The obtained instance segmentation for different microbial colony types correctly reproduces the colony shapes.
%However, a problem with achieving exact segmentation masks appears at the patch corners giving not fully circle-shape colonies. This problem can be mitigated by adding appropriate padding to the evaluated patches, segmenting a given fragment several times (using patches that overlap), and then performing prepossessing by proper joining such redundant masks (a similar method with redundant bounding boxes was used for the detection task).

\subsubsection*{Colony counting metrics}
One of the main applications of object detection in microbiology is to automate the process of counting microbial colonies grown on a Petri dish. 
We verify the proposed method of synthetic dataset generation by comparing it with a standard approach where we collect a big real dataset~\cite{agar} and train the detector for colony detecting and counting tasks.

There are several types of metrics typically used in the context of object detection and counting. 
To characterize the quality of detection, we calculate the standard mean Average Precision (mAP) established by the famous COCO competition\cite{coco}. The Average Precision (AP) of detection is calculated for a given Intersection over Union (IoU) threshold. IoU describes the level of overlapping between the truth and predicted bounding boxes. The mean value of AP for different IoU thresholds and classes gives the mAP. 
To measure the effectiveness of colony counting, two separate metrics were used, namely, standard Mean Absolute Error (MAE) and symmetric Mean Absolute Percentage Error (sMAPE), which additionally weights the errors using information about the number of instances present on a given dish. Precise definitions of both measures can be found in Supplementary Material for [\citeonline{agar}].

\begin{figure}[bt]
\centering
\includegraphics[width=.3\linewidth]{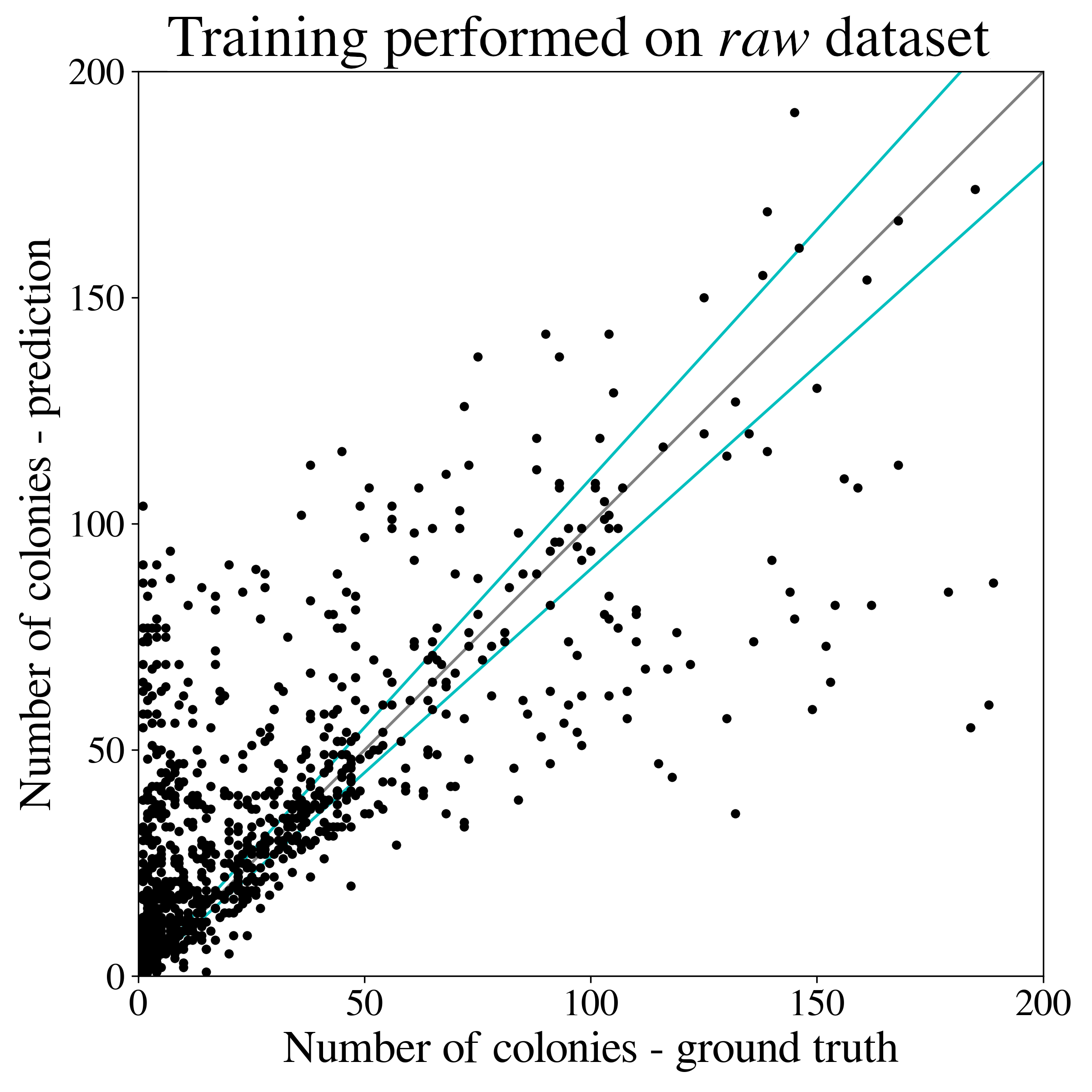}
\includegraphics[width=.3\linewidth]{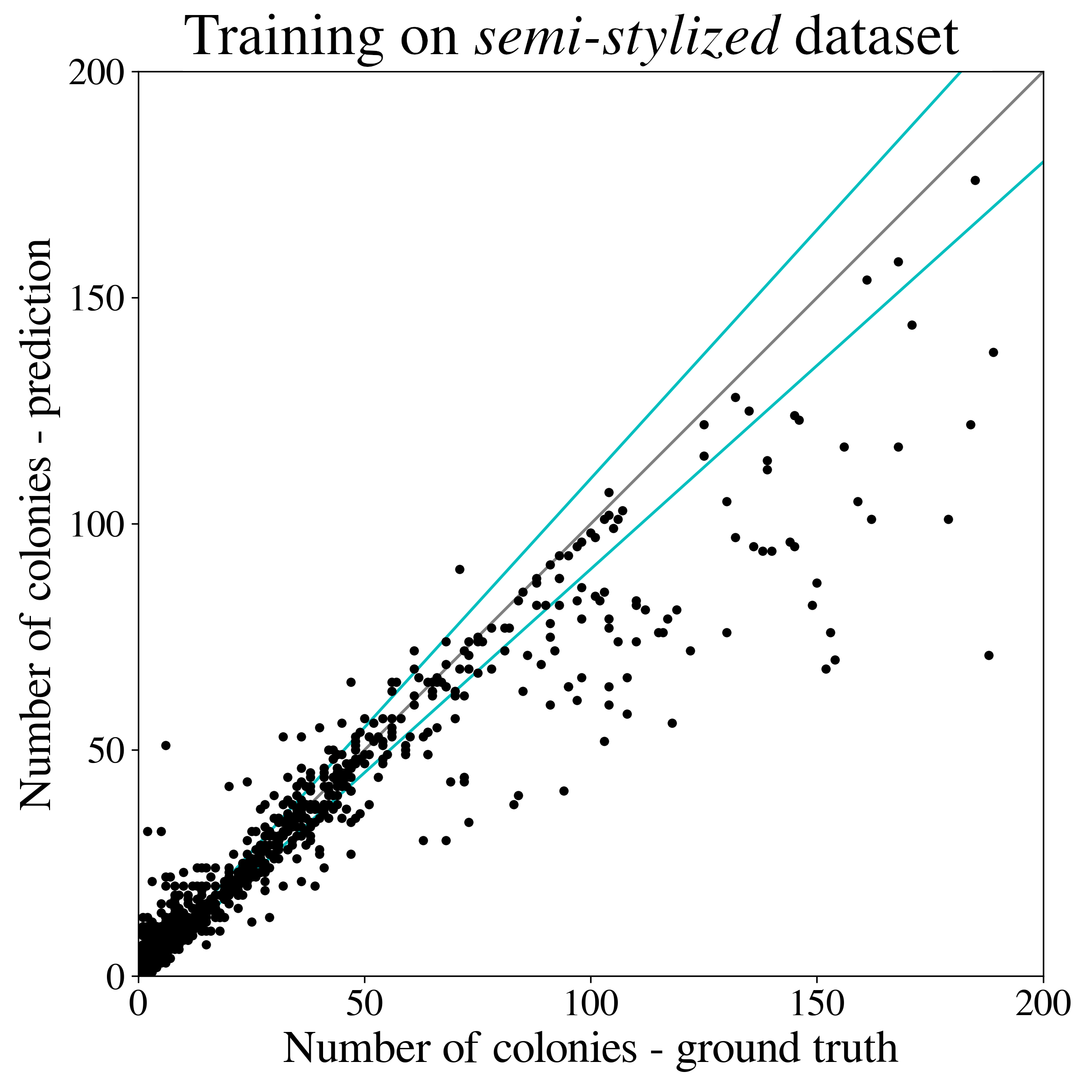}
\includegraphics[width=.3\linewidth]{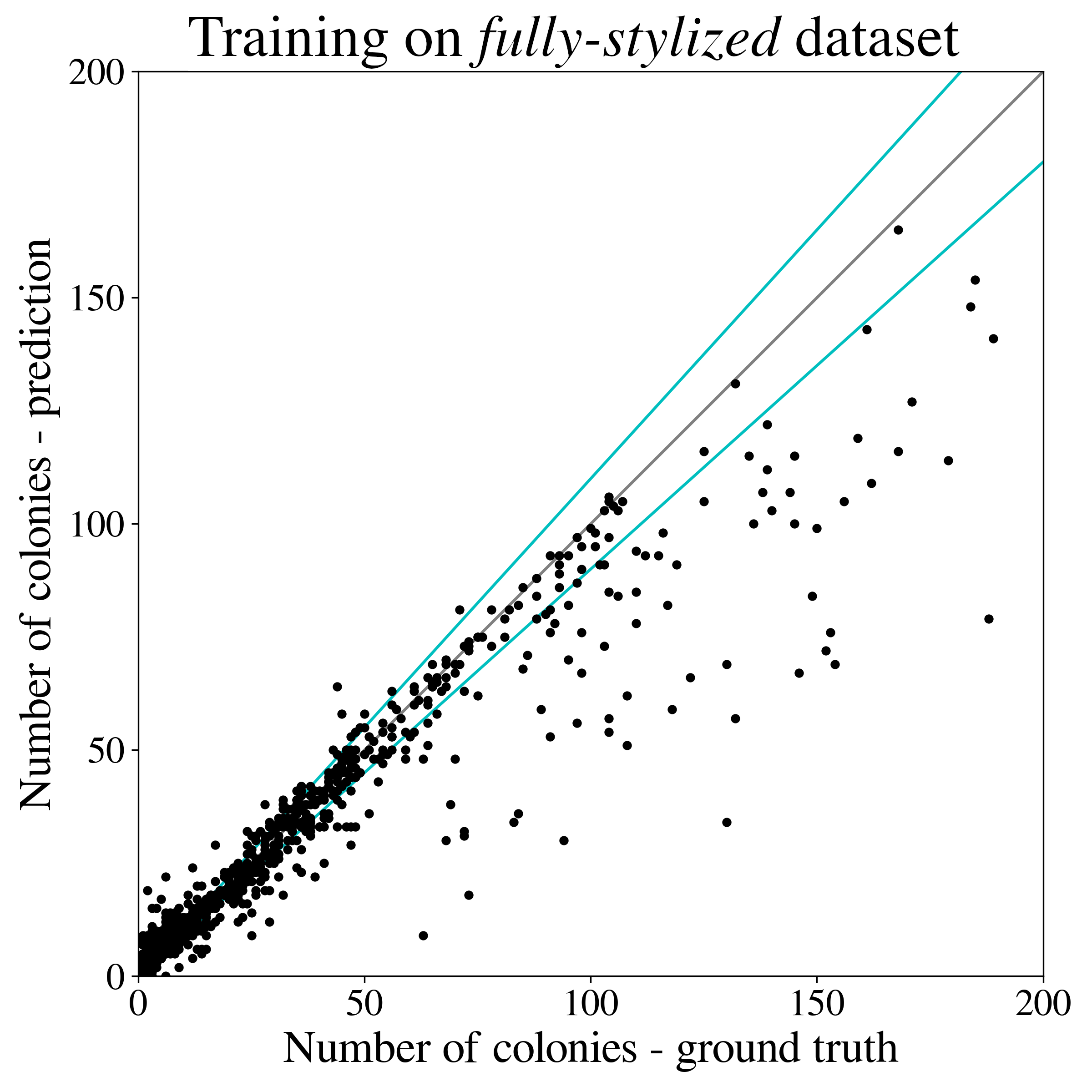}\\
\includegraphics[width=.3\linewidth]{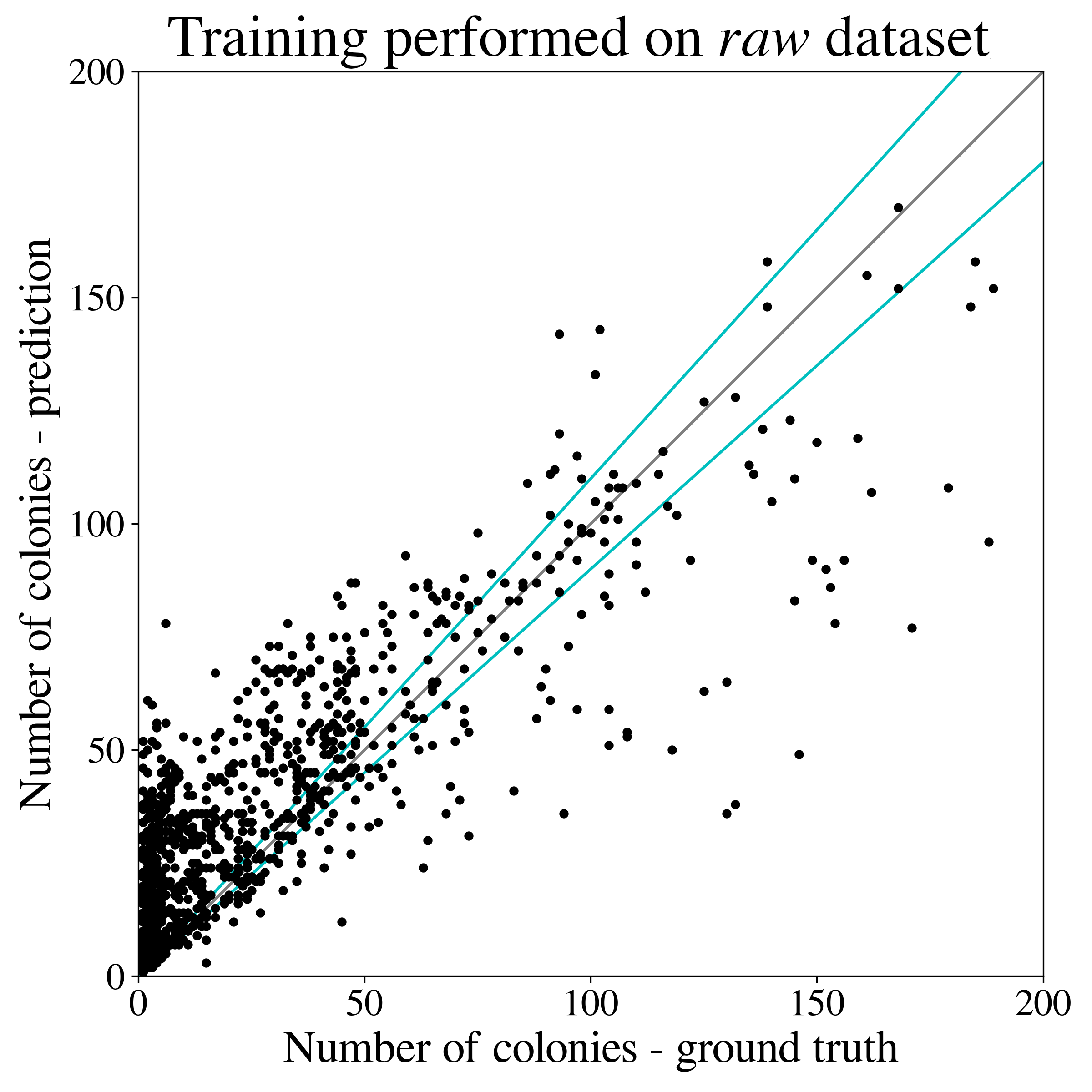}
\includegraphics[width=.3\linewidth]{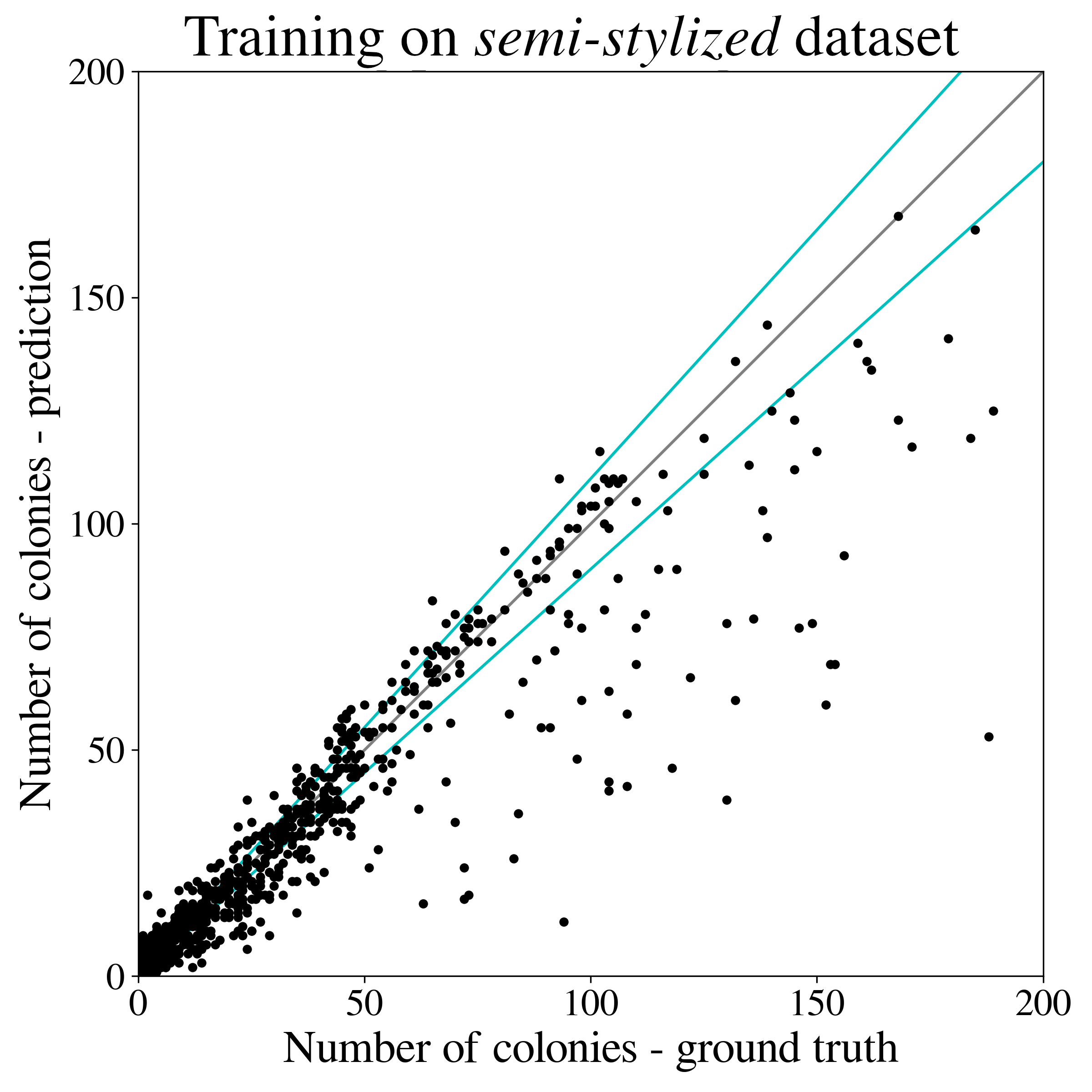}
\includegraphics[width=.3\linewidth]{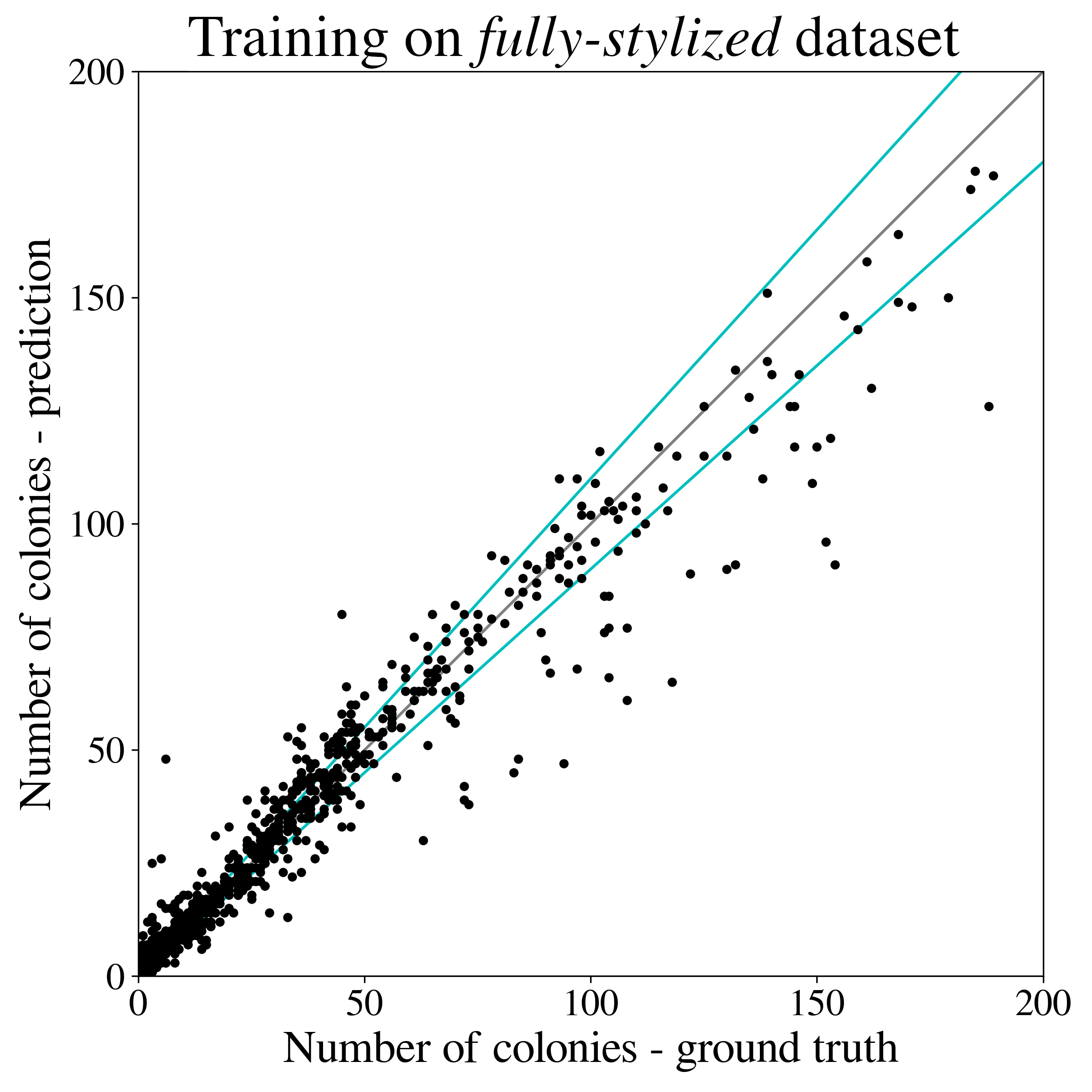}
\caption{Microbial colonies counting tested on real data for three different synthetic training datasets: \textit{raw} (left), \textit{semi-stylized} (middle), and \textit{fully-stylized} (right). In ideal detection every black dot representing a single Petri dish image should lay on the $y=x$ black line. Top row shows results for Faster, while bottom row for the Cascade detector.}
\label{fig:count}
\end{figure}

Let us now discuss the results of counting microbial colonies on a Petri dish. We train the chosen detector (Faster RCNN with ResNet-50 backbone and Cascade R-CNN with HRNet) on a 50k large dataset generated using 100 images of the training part of the \textit{higher-resolution} AGAR subset, and test microbial colonies counting in the same task as conducted in [\citeonline{agar}] (testing part of AGAR \textit{higher-resolution}). 
The resulting  detection metrics are presented in Table~\ref{tab:res}, and counting results in Fig.~\ref{fig:count}
(the detectors were trained on \textit{raw}, \textit{semi-stylized}, and \textit{fully-stylized} sets separately).
\begin{table}[htb]
\centering
\begin{tabular}{|l|c|c|c|c|c|c|}
%hline
\multicolumn{1}{c}{} &  \multicolumn{3}{c|}{Faster R-CNN} &  \multicolumn{3}{c}{Cascade R-CNN}\\
\hline
\multicolumn{1}{|r|}{metrics} & \cellcolor[HTML]{EFEFEF}detection & \multicolumn{2}{c|}{\cellcolor[HTML]{EFEFEF}counting} & \cellcolor[HTML]{EFEFEF}detection & \multicolumn{2}{c|}{\cellcolor[HTML]{EFEFEF}counting}\\ 
\hline
datasets & \cellcolor[HTML]{EFEFEF}mAP (IoU=0.50:0.95) & \cellcolor[HTML]{EFEFEF}MAE & \cellcolor[HTML]{EFEFEF}sMAPE & \cellcolor[HTML]{EFEFEF}mAP (IoU=0.50:0.95) & \cellcolor[HTML]{EFEFEF}MAE & \cellcolor[HTML]{EFEFEF}sMAPE \\ \hline
\cellcolor[HTML]{EFEFEF}\textit{raw}  & 
0.203  & 17.54 & 33.02\% & 0.201 & 17.89 & 34.38\% \\ \hline
\cellcolor[HTML]{EFEFEF}\textit{semi-stylized} & 0.329 & 6.26 & 12.91\% & 0.325 & 6.62 & 15.35\% \\ \hline
\cellcolor[HTML]{EFEFEF}\textit{fully-stylized} & 0.401 & 5.82 & 11.65\% & 0.416 & 4.49 & 10.83\% \\ \hline\hline
\cellcolor[HTML]{EFEFEF}real [\citeonline{agar}] & 0.493
& 4.75                   
& 5.32\%
& 0.520
& 4.31
& 4.86\%
\\ \hline
\end{tabular}
\caption{\label{tab:res} Microbial colonies detection tested on real data for three different synthetic training datasets: \textit{raw}, \textit{semi-stylized}, and \textit{fully-stylized}. For comparison, we also provide results for training on the real dataset\cite{agar}. We present results for Faster R-CNN and Cascade R-CNN detectors along with various metrics describing detection (mAP), and counting fidelity (MAE, sMAPE).}
\end{table}
% 33.02 (95 5) 
% 12.91 (80 7)
% 11.65 (70 7)
It turns out that the detection precision itself (mAP) and counting errors (MAE and sMAPE) for the \textit{fully-stylized} set are only slightly worse (especially MAE) than for the same detector but trained on the whole big set containing over 7360 of real images giving about 65k of patches. 
%Results for training on the synthetic data are stored in three upper rows, while for training on the real data are stored in the last row in Table~\ref{tab:res}.
Cascade R-CNN detects colonies slightly better than Faster R-CNN and this is true for training on both synthetic and real data. It is also interesting to note that Cascade R-CNN is better for the most stylized set, for the others, \textit{semi-stylized} and \textit{raw}, Faster R-CNN performs better.
%$\mathrm{mAP}=0.493$, $\mathrm{MAE}=4.75$, $\mathrm{sMAPE}=5.32\%$
%$\mathrm{mAP}=0.520$, $\mathrm{MAE}=4.31$, $\mathrm{sMAPE}=4.86\%$

It is also clear that introducing style transfer augmentation greatly improves the detection quality, and without stylization the results are rather poor---see results for \textit{raw} set in Fig.~\ref{fig:count}(left).
The top row in Fig.~\ref{fig:count} shows counting results for Faster, while the bottom row for Cascade detector.
Moreover, in the Fig.~\ref{fig:count}(right)---results for \textit{fully--stylized} set, we observe the typical problem where the detector underestimates the number of colonies for highly populated dishes. This is because in such cases colonies (especially bigger ones) frequently overlap, which makes detection (and counting) harder.

\section*{Discussion}
Object counting problems, such as estimating the number of cells in a microscopic image or the number of humans or cars in a video frame, have become one of the major tasks in computer vision these days\cite{dis1}.
Generation of synthetic microbiological images using simple geometrical transformations that emulate microscopic views of bacterial colonies was used for validation of various algorithms for automated image analysis\cite{dis2}, or to train a high-resolution convolutional neural network to build density-map-based cell counters\cite{dis3}. In the latter, authors use methods similar to ours, where the model was trained using synthetic data and tested on real images. However, in our approach we train a much more complicated but thus more flexible (and prone to generalize) object detector. Therefore, we need to use a much bigger and more diverse dataset containing objects of different sizes and with different lighting conditions. We also verified that further enlarging the dataset by adding another 10k patches practically did not improve the detection fidelity.
Deep CNNs were also trained to count cells\cite{dis4} and microbial colonies\cite{ferr,dnn}, but in these approaches the network training was done entirely in a supervised manner using real datasets.
%also Synthetic Data and Artificial Neural Networks was used for Natural Scene Text Recognition\cite{dis5}

Simple augmentation techniques were applied successfully to extend microbial datasets\cite{dis6,ferr} in bacteria classification tasks. We also flipped or rotated the images, or added salt and pepper noise or speckle noise, but with no significant effect on the training process. According to our experiments, only advanced augmentation by using style transfer significantly affects the training of the detector. It is also worth mentioning that adding more styles did not help much; during the generation we now choose from 20 style images, but when we increased this number to 50, it did not give further improvement. On the other hand, the degree of styling was more important, but if it was too high, i.e $\lambda>0.05$, the fidelity of microbe classification would decrease.
Other groups involved in the generation of synthetic microbial images also reported the usefulness of style transfer methods.\cite{bib:Andreini2020}
They use the style transfer algorithm in the microbial colony segmentation task to improve the realism of synthetic images of Petri dishes generated using GANs. However, we additionally show that style transfer allows us not only to improve realism, but also to introduce different color or lighting domains, and thus enhance the trained detector's ability to generalize.
% jakas glupote napisalem ale troche nas zabija ta praca 
We saw in Fig.~\ref{fig:det} that by using different style images during training, the detector can learn to count in slightly different lighting conditions (domains) at the same time. Therefore, our approach can be used to quickly change domains, e.g., we can collect styles and empty dishes in different lighting, and then generate synthetic images in that domain.

It is also worth emphasizing that standard GANs are not applicable in our case because to train the detector in a supervised manner we need to know exactly where each colony will appear. We need to control precisely where they are located---their placement and number on a dish, which is not possible with generic, unconstrained GAN. On the other hand, there are variations of GANs such as cycle GAN or conditional GAN (e.g. pix2pix translation~\cite{congan}) that would allow colonies to be generated at desired locations. However, our preliminary experiments have shown that while we are able to generate realistic looking images in a single style with pix2pix, training of this type of networks is difficult and it is much harder to achieve sufficient diversity of domains using this approach. As a result, the detector trained on such data performs worse. On the other hand, one-shot style transfer is simply faster (especially for high resolution images), easier to train (without mode collapse or converge failure), and able to transfer with high resolution to different domains. Moreover, data leakage, which is more problematic for medical investigation due to privacy issues, can be controlled with proposed generation strategy. Therefore, to generate the data we used a hybrid of traditional computer vision and image style transfer rather than individual generative models.

\section*{Conclusions}
% musimy jeszcze podsumować całość
Training modern deep learning models requires large and diverse datasets containing thousands of labeled images\cite{coco}. By using traditional computer vision techniques complemented by a deep learning style transfer algorithm, we were able to build a microbial data generator supplied with only 100 real images that demands much less effort and resources than collecting and labeling a vast dataset containing thousands of real images.

In principle, any object with some local differences from the background (in colorspace or brightness) can be segmented and then used to generate images in our scheme together with the appropriate transfer of styles. We showed that, once generated, our synthetic dataset allows us to train state-of-the-art deep learning object detectors,
making it applicable to any object detection task in microbiology, but also in other scientific or industrial applications.

\section*{Acknowledgements}
Authors would like to thank Agnieszka Pawlak for valuable discussions.
Project “Development of a new method for detection and identifying bacterial colonies using artificial neural networks and machine learning algorithms” is co-financed from European Union funds under the European Regional Development Funds as part of the Smart Growth Operational Program. Project implemented as part of the National Centre for Research and Development: Fast Track (grant no. POIR.01.01.01-00-0040/18).

\bibliography{main}

\end{document}